\documentclass[10pt]{extarticle}
\usepackage{microtype}
\usepackage{spconf,amsmath}
\usepackage[final]{graphicx}
\usepackage{url}
\usepackage[moderate]{savetrees}
\usepackage{xcolor}
\usepackage{color} 	
\newcommand{\ignore}[1]{}

\title{Using of heterogeneous corpora for training of an ASR system}
\name{Jan Trmal$^{1,2}$, Gaurav Kumar$^{1,2}$, Vimal Manohar$^{1}$, Sanjeev Khudanpur$^{1,2}$, Matt Post$^{2}$, Paul McNamee$^{2}$\thanks{The work was supported by the NSF CRI Grant No 1513128, by the DARPA LORELEI Contract No HR0011-15-2-0024 and by IARPA BABEL Contract No 2012-12050800010. "The U.S. Government is authorized to reproduce and distribute reprints for Governmental purposes notwithstanding any copyright annotation hereon.}}
\address{$^1$Center for Language and Speech Processing\\
$^2$Human Language Technology Center of Excellence\\
The Johns Hopkins University, Baltimore, MD 21218, USA}

\begin{document}
%\ninept
%
\maketitle
\begin{abstract}
The paper summarizes the development of the LVCSR system built as a part of the Pashto speech-translation system at the SCALE (Summer Camp for Applied Language Exploration) 2015 workshop on ``Speech-to-text-translation for low-resource languages''.
The Pashto language was chosen as a good ``proxy'' low-resource language, exhibiting multiple phenomena which make the speech-recognition and  and speech-to-text-translation systems development hard.

Even when the amount of data is seemingly sufficient, given the fact that the data originates from multiple sources, the preliminary experiments reveal that there is little to no benefit in merging (concatenating) the corpora and more elaborate ways of making use of all of the data must be worked out.

This paper concentrates only on the LVCSR part and presents a range of different techniques that were found to be useful in order to benefit from multiple different corpora

\end{abstract}
\begin{keywords}
speech translation; pashto; babel; multiple corpora; neural networks;  discriminative training;
\end{keywords}
\section{Introduction}
\label{sec:intro}
Pashto belongs to the southeastern Iranian branch of Indo-Iranian languages. It has
three main variants: Northern and Central (both spoken mainly in Pakistan)
and Southern (spoken mainly in Afghanistan). Each of these variants has a number
of dialectal varieties. It is estimated that Pashto has 66 million speakers
across the world\cite{ethnologue}.
While written Pashto has existed since the 16th century, standardization of the writing
system is still in progress. There are a substantial number of words that have more
than one publicly accepted way of being written (cf. English adviser
vs.~advisor). Other issues include present/missing spacing after certain characters (especially
those, which belong to non-connecting arabic characters, and frequent substitution
of visually similar (and similar sounding) characters.
The last problem is emphasized by the fact that for writing Pashto, several different
keyboard layouts are used ``in the wild''. There is, of course the official Pashto
keyboard layout although the Arabic and Urdu layouts are used as well.  The alternative
layouts have a majority of the Pashto characters (and people freely substitute those
which are missing with visually similar characters).
Also, different fonts can have small deficiencies in rendering of glyphs, especially
during kerning or joining of characters and the users often try to fix this by
substituting a different character that looks better (i.e. closer to the expected shape)
in the given context.

\begin{figure}
  \centering
  \includegraphics[width=1cm]{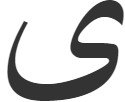}
  \hspace{1cm}
  \includegraphics[width=1cm]{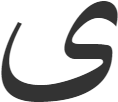}
  \caption{\textsc{Arabic Letter Farsi Yeh}
and \textsc{Arabic Letter Alef Maksura}}\label{fig:glyphs}
\end{figure}

A direct impact of this is that a word as a sequence of glyphs
(visual representations of characters) can be represented as multiple sequences
of unicode codepoints (numerical codes of the characters).
One example was the pair of unicode codepoints
\textsc{Arabic Letter Farsi Yeh} and \textsc{Arabic Letter Alef Maksura} (see Figure~\ref{fig:glyphs})
whose glyphs are rendered with no visible differences, despite the fact that the
codepoints are different.

This also causes problems when looking only for visual differences (for example, during
debugging of problems with the lexicon). We mention this somewhat anecdotal
evidence to show the readers who are only familiar with languages which use romanized
scripts, that there are a number of hidden peculiarities that are really surprising
when met for the first time.

\section{Speech Corpora available}
In this section, we give a short overview of the different corpora we worked with.  We acquired two
corpora before the start of the workshop and one additional during the course of the workshop.
As the latter is substantially different and not completely suitable for our task (LVCSR and machine
translation), we did not achieve any gain from using it and we are reporting the numbers here just for
the sake of completeness.%\vspace{-1.0em}

\subsection{Lila consortium Appen Pashto (A)}
The Appen Pashto (dataset ``A'') contains approximately 90 hours of conversational telephone (8\,kHz)
speech. As the train/dev partitions were not externally defined in the Appen data release, we  partitioned the data
into 85 hours for training and 5 hours as dev sets\footnote{These splits are part of the Kaldi recipe}.
The lexicon was included with the corpus and the lexical entries included vowelized representations and the
romanized forms of words. Moreover, the lexicon contained four dialectal variants of
pronunciation of each word (we assume that these were generated automatically).\vspace{-1.0em}

\subsection{IARPA Babel Pashto (B)}
Babel Pashto (dataset ``B'') is the Full~Language Pack\footnote{\texttt{release-IARPA-babel104b-v04.bY}} in Babel program terminology.
Simply explained, it is a dataset of 80 hours of 8\,kHz sampled telephone speech data and the associated lexicon and transcripts.
We used the development set defined in the corpus. Although our description of the corpora ``A'' and ``B'' may indicate that the datasets
are similar, our observations demonstrated that this was not the case. We observed that dataset ``B'' was, on overall, prepared and transcribed more carefully than dataset ``A''.\vspace{-1.0em}

\subsection{TransTac Pashto (T)}
The TransTac Pashto is significantly different when compared with the two aforementioned corpora.
By nature, it is more scripted speech, albeit with high level of spontaneity.
It is also professionally recorded (in a recording room) and hence we had to down-sample it to 8\,kHz
for use with corpora ``A'' and ``B''.
The dataset was created as follows. The participating individuals were given
one out of a limited set of scenarios and they were then asked to re-enact that scenario.
We were unable to obtain the definitions of the train/dev partitions for this dataset
that were used in the DARPA TransTac (\cite{sanders2013evaluation}) project.
Because of this and the fact that we used the down-sampled speech, our performance
is not directly comparable to the previously reported results for this dataset.
We started work on this corpus relatively late in the course of the workshop and
hence a majority of the reported experiments done on the ``merged'' training data only
used the A+B training set. Also, our primary aim was not to achieve the best result on
this corpus but to establish whether this dataset could be used in the 8\,kHz telephone speech scenario.\vspace{-1.0em}

\subsection{Corpus preparation}
\label{ssec:corpora-prep}
As mentioned in the introduction, there is significant variability in the process used to transcribe
corpora ``A'' and ``B''. Our first efforts targeted at making the transcriptions from both corpora more
consistent in their use of characters. This was motivated by our inspection of the lexicons from the corpora.
Starting with the lexicons and then removing the vowel marks from all words, the overlap (defined as the number of words the lexicons share) was only 15\,\%. In an attempt to determine the minimal number of character
changes that would increase lexicon overlap, we developed a simple algorithm that allowed a language expert to determine the~edit~rules.

\begin{enumerate}
  \setlength\itemsep{-0.00em}
  \item Take 1000 most frequent words from each lexicon (based on the associated transcripts). We chose this number
  so as to cover over 90\,\% of the frequency mass in each corpus.
  \item Find the best word pairs in terms of the least character distance, one word from each lexicon, based on
  \begin{enumerate}
    \item Character Edit Distance
    \item Phone Edit Distance: We used the \emph{Festvox toolkit}\footnote{\texttt{http://festvox.org}} to generate the phone sequence for each word in the lexicon.
  \end{enumerate}
  \item Generate count based statistics for character based substitution and deletion rules based on the previous step.
  \item Get an expert to verify the most frequent rules.
  \item Use the rules to modify the lexicons and transcripts.
  \item Go to Step 1.
\end{enumerate}

To elaborate, step 2 (the edit character computations) can provide highly informative
observations -- for each edit operation, we keep track of the global effect of applying such rules. We choose the minimal set of rules that would lead to maximal corpus overlap. These most frequent operations can provide insights to discover character-substitution or deletion rules. See Table~\ref{table:rules} for an example of these
automatically discovered (and expert-confirmed) rules.
\begin{table}
  \centering
  {\tiny
  \begin{tabular}{|c|l|l|}
    \hline
    Operation & Character & Character \\
    \hline
    DEL & \textsc{Arabic Kasra} &  \\
    DEL & \textsc{Arabic Fathatan} &  \\
    DEL & \textsc{Arabic Kasratan} & \\
    DEL & \textsc{Zero Width Non-Joiner} &\\
    SUB & \textsc{Arabic Letter Kaf} & \textsc{Arabic Letter Keheh} \\
    SUB & \textsc{Arabic Letter Gaf} & \textsc{Arabic Letter Kaf With Ring}\\
    SUB & \textsc{Arabic Letter Farsi Yeh} & \textsc{Arabic Letter Yeh}\\
    SUB & \textsc{Arabic Letter Yeh With Tail} & \textsc{Arabic Letter Yeh} \\
    SUB & \textsc{Arabic Letter Yeh With Hamza Above} & \textsc{Arabic Letter Yeh}\\
    SUB & \textsc{Arabic Letter E} & \textsc{Arabic Letter Yeh}\\
    SUB & \textsc{Arabic Letter Alef With Hamza Above} & \textsc{Arabic Letter Alef}\\
    \hline
  \end{tabular}
  }
  \caption{Examples of automatically discovered rules for lexicon/text normalization, confirmed later by a language expert. DEL = Deletion, SUB = Substitution`'}\label{table:rules}
\end{table}
After about four iteration of this process, we increased the overlap of the most frequent 1000 words across corpora ``A'' and ``B'' to approximately 70\,\%. After the fourth iteration, we didn't find any additional systematic differences, so we stopped there.

\section{Baseline acoustic system}
\label{sec:baseline}
\begin{table}
  \small
  \centering
  \begin{tabular}{|c||c|c||c|}
    \hline
        & \multicolumn{3}{c|}{dev set WER}  \\
        & A-dev & B-dev & T-dev\\
    \hline
    triphone GMM system  & 64.10\,\% & 61.90\,\% & 36.45\,\%\\
    $+$ pron. probs      & 62.03\,\% & 60.30\,\% & 35.37\,\%\\
    $+$ TDNN system      & 52.09\,\% & 48.19\,\% & 25.29\,\%\\
    \rule[-1.2ex]{0pt}{2.5ex}
    $+$sequence training & 48.33\,\% & 45.43\,\% & 23.23\,\%\\
    \hline
    \rule{0pt}{2.3ex}
    $+$duration modeling & 48.19\,\% & 44.93\,\% & 23.12\,\%\\
    \hline
  \end{tabular}
  \caption{Performance of the baseline ASR system on three Pashto test sets: ``A'' - the Appen Lila corpus, ``B'' - the Babel corpus, and ``T'' - DARPA TransTac corpus.}\label{table:overall_baseline}
\end{table}
In this section, we provide a high level description of how we trained the baseline system.
As the objective was to develop a sufficiently simple, single pass, minimum delay system, we opted for training
a deep neural network system (or more precisely, a TDNN, as described in \cite{peddinti2015multisplice}).
We used the Kaldi toolkit \cite{Povey_ASRU2011} for training the ASR system. There are a couple of details worth mentioning which follow.

\subsection{Speed perturbation}
We utilized augmentation of the data via speed perturbation (as described in \cite{ko2015augmentation})
during training. We used \emph{sox}\footnote{\url{http://sox.sourceforge.net/}} to obtain two copies of the training data (the first copy used a speed factor $1.1$ and the second $0.9$). Our experience confirms this improves overall robustness of the resulting models.
The network was trained in a parallel fashion using model averaging, as described in \cite{povey14parallel}.

\subsection{Estimation of pronunciation probabilities}
The pronunciation lexicon rarely contains the probabilities of the individual
pronunciation variants. It is however possible to estimate these probabilities from
the alignments of the training data.  Moreover, it is possible to model word-dependent
silence probabilities, in addition to modeling of the probability of
silence to estimate (and suitably smooth) the probability of each word appearing
after silence. See \cite{chen2015pronunciation} for a detailed analysis
of this and several related ideas.

In our experience, it is also beneficial to re-estimate these probabilities iteratively
several times during the training process. We saw reasonable gains (given the fact
that this phase itself is not computationally expensive) from using the probabilistic
pronunciation lexicon even during training.

\subsection{Sequence discriminative training}
For the sequence training  we used the sMBR method \cite{kaiser2000novel,gibson2006hypothesis} which
is reported \cite{vesely2013sequence} to give best performance (measured with respect to WER).
Also, we found it beneficial to adjust the prior probabilities -- used during decoding for converting TDNN posterior probabilites into likelihoods - after finishing the discriminative training.

Historically, the priors are computed from alignment. As mentioned in \cite{manohar2015semi},
marginalizing of the DNN posteriors over all acoustic vectors gives better performance, especially
when the data is noisy. Only a limited subset of the full training data is usually needed,
so again, this improvement comes relatively cheap.

\subsection{Duration model rescoring}
After the final lattices were generated, we used the duration modeling rescoring as described
in \cite{alumae2014neural}. We used the software the author of the paper provided.\footnote{\url{https://github.com/alumae/kaldi-nnet-dur-model}}
We find the improvements fairly consistent, albeit lower, than the numbers reported in the original paper.

\subsection{Overall performance of the baseline system}
The overall performance of the resulting baseline system is reported in Table~\ref{table:overall_baseline}.
Please note that a separate baseline system is trained for each test set -- Appen, Babel and TransTac -- using only speech data from the corresponding training set.

\section{Related work}
\begin{table}
  \small
  \centering
  \begin{tabular}{|l|c|}
    \hline
    \multicolumn{2}{|c|}{Babel FullLP (B-dev) performance} \\
    \hline\hline
    Radical-JHU 3-way combination             & 50.70\,\%\\
    Radical-JHU single system best number     & 53.60\,\%\\
    \hline
    Scale B-train only without normalization  & 47.30\,\%\\
    Scale B-train only with normalization     & 43.92\,\%\\
    Scale A+B-train with normalization        & 45.14\,\%\\
  \hline
  \end{tabular}
  \caption{Comparison of the performance (WER) of the newly developed baseline vs. the best single system previously developed by us for the Babel dataset.}\label{table:comparison}
\end{table}
This comparison is not completely straightforward.
There was a substantial amount of work reported as a part of the Babel Pashto project, but the reported numbers are generally from a combination of multiple (sometimes of huge number of) systems.
As our team participated (as a part of team Radical) in this project, we are presenting a comparison of our current system with respect
to our best-performing Babel system (hybrid DNN system) from two years ago (see Table~\ref{table:comparison}).
The comparison has to be made carefully, as we (in order to unify the corpora at hand) applied the rule set mentioned in section~\ref{ssec:corpora-prep}. From Table~\ref{table:comparison} it can be seen that the new training procedure (Scale B-train only without normalization) gives
us $\sim 6$\,\% absolute gain over the older system and the normalization rules provide another $\sim 3$\,\% gain.
Note, as an aside, that simply adding the Appen data to the Babel trainng set degrades WER by about 1.2\,\%.

\section{Joint Multi-corpus Training}
\label{sec:joint-multi}
During the course of the workshop, it became apparent that the three corpora actually do not
combine well. The corpora ``A'' and ``B'' are closest, but even their combination
for training did not produce better results -- see Table~\ref{table:mixing_data}.

\begin{table}
  \small
  \centering
  \begin{tabular}{|c||c|c|c|}
    \hline
                          & A-dev     & B-dev     & T-dev     \\
    \hline\hline
    ``native'' data       & 49.49\,\% & 43.92\,\% & 23.12\,\% \\
    A+B-train             & 48.19\,\% & 45.14\,\% & 53.46\,\% \\
    \hline
  \end{tabular}
  \caption{Comparison of the performance of the system trained on corpus-specific(native) data vs the training set obtained by merging the training data of ``A'' and ``B'' sets (TDNN system)}\label{table:mixing_data}
\end{table}

Another piece of evidence can be gathered from Table~\ref{table:mixing_lms}. The language model created from the training data of the~``T''~dataset was not useful for the language model interpolation.
\begin{table}
  \small
  \centering
  \begin{tabular}{|c||c|c|c||c|c|}
\hline
       & lm A   & lm B & lm T & ppl orig & ppl interp\\
\hline\hline
text A & 0.8    & 0.2  & 0.0  & 99.2  & 92.9\\
text B & 0.1    & 0.8  & 0.1  & 141.9 & 140.0\\
text T & 0.0    & 0.0  & 1.0  & 86.7  & 86.7\\
\hline
  \end{tabular}
  \caption{Optimal mixing weights, and the resulting perplexities on three dev-test sets, for interpolating language models trained on the three corpora: ``A'' -- Appen, ``B'' -- Babel and ``T'' -- TransTac.}\label{table:mixing_lms}
\end{table}

\begin{figure}
  \centering
  \includegraphics[width=0.44\textwidth]{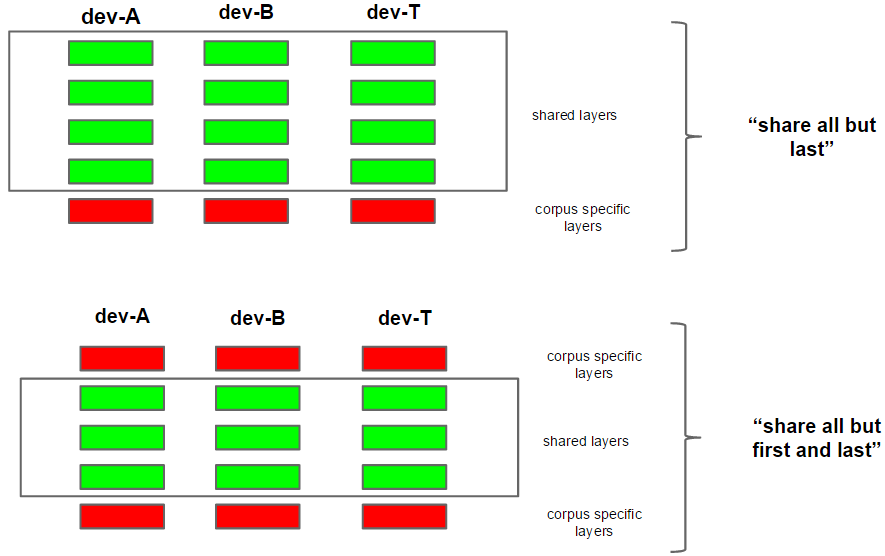}
  \caption{Scheme of sharing the layers for joint multi-corpora training. The best performance
  was achieved with the ``share all but first and last'' approach.}\label{fig:multicorpora}
\end{figure}

As the diversity of the data proved to be too high to allow for the training of a single model on all of the data that would function well, we decided to train data-set specific models, i.e. train three models, each of which would be specialized to that given dataset.
Moreover, we tried to find out if there was a way to benefit from the fact that we had multiple (similar) corpora. The method we used to exploit this fact was the sharing (i.e. training jointly) of the hidden layers and only having the last and the first layers be dataset-specific. The reason for doing this was two-fold. First, it allowed us to train a larger neural network with potentially better performance. Secondly, the shared layers would hopefully learn more general/robust hyperplane separations. See Figure~\ref{fig:multicorpora} for an illustration of this method.

We experimented with different sharing strategies. The best performance was obtained when the first and the last layer were shared.
Sharing less or more layers (than the first and last one) has made the performance worse.
\begin{table}
  \small
  \centering
  \begin{tabular}{|c|c|c|c|}
    \hline
    \rule{0pt}{2.3ex}
                                                & dev-A      & dev-B     & dev-T \\
    \hline\hline
    \rule{0pt}{2.3ex}
    "Matched" single corpus                             & 55.38\,\% & 46.81\,\% & 24.76\,\%\\
    A+B-train                                   & 51.89\,\% & 48.19\,\% & 52.59\,\%\\
    \hline
    \rule{0pt}{2.3ex}
    share all but last                          & 53.20\,\% & 47.35\,\% & 27.03\,\%\\
    \rule[-1.2ex]{0pt}{2.5ex}
    share all but first and last                & 51.22\,\% & 45.02\,\% & 25.38\,\%\\
    \hline
    \rule{0pt}{2.3ex}
    $+$ optimized LM                            & 50.83\,\% & 44.77\,\% & 25.38\,\%\\
    $+$ duration modeling                       & 50.44\,\% & 44.26\,\% & 24.83\,\%\\
    \hline
  \end{tabular}
  \caption{ASR performance (WER) for multi-corpus training of acoustic models, with shared hidden layers and corpus-specific input and/or output layers.}\label{table:multilang}
\end{table}

The training procedure was similar to the training of a single network. First, in each step, for each dataset, a new, updated network was obtained using model averaging. After that, these corpus-specific networks were averaged and the shared layers of this final network were copied back to the corpus-specific network. This represents one iteration of the joint-multi-corpora training.

Using the described approach, we were able to train a 7-layer p-norm network with a hidden layer dimension of 4500 and p-norm pooling of 1$:$10. The best network trainable using only the native data was a 7-layer p-norm network with a hidden layer dimension of 2500 and p-norm pooling of 1$:$10.

\section{Main Results}
A key lesson from Table~\ref{table:multilang} is that relative to single corpus training, straightforward pooling of multiple corpora (A and B) to train a single acoustic model results in \emph{significant degradation} of WER on 2 out of 3 test sets (Babel and TransTac).  Only the Appen test set benefits from training on the Babel data.

The main new result, by contrast, is that training separate TDNN acoustic models for each corpus while sharing the internal layers -- an idea akin to the training of multilingual acoustic models -- followed by some LM and duration model optimization results in significant improvements in WER.

The Appen WER reduces from 51.9\,\% to 50.4\,\% (3\,\% relative), while the Babel WER reduces from 46.8\,\% to 44.3\,\% (5\,\% relative).

The TransTac corpus is too different from the Appen and Babel corpora to see any benefit from data pooling.  However, it is also remarkable that acoustic models trained on only the Appen and Babel data with the \emph{new method} attain the same performance as models trained exclusively on TransTac data: 24.8\,\%.  By contrast, acoustic models trained on Appen and Babel using traditional data pooling degrade WER to 53\,\%.  This illustrates the cross-corpus robustness of the new method.

\section{Conclusion}
This paper presents an overview of Pashto low-resource ASR system built during the SCALE'15 workshop.
Initially, we developed a single pass LVCSR system for Pashto language. This system, trained using a corpora obtained by concatenating two different corpora (called ``A'' and ``B'' in this paper) outperformed significantly the performance of systems developed as a part of the Babel program. Our single-pass system achieved comparable, if not better, performance with respect to very complex systems (multiple passes + combination of multiple systems).

While achieving as good WER as possible was important, the body of the work done during the duration of the workshop and hence described in the paper concentrates on providing insight on how to combine training data for multiple different sources.  Pashto is a good proxy to demonstrate various issues which can be seen ``in the wild''.  While the aspects of having different audio channels (i.e. sampling frequencies, additive line noise, room impulse response) are generally fully appreciated, the issues stemming from an inner complexities of the language are largely overlooked.  On Pashto, we are demonstrating some issues native to languages with not mature-enough computer-implemented writing system together with our take on how to deal with them in order to obtain data homogeneous enough to be useful for ASR system training.

\vfill\pagebreak

\label{sec:refs}

\bibliographystyle{IEEEbib}
\bibliography{refs}

\end{document}